\documentclass{article}

\PassOptionsToPackage{numbers, compress}{natbib}
 \usepackage[preprint]{neurips_2026}

\usepackage[utf8]{inputenc} 
\usepackage[T1]{fontenc}    
\usepackage{hyperref}       
\usepackage{url}            
\usepackage{booktabs}       
\usepackage{amsfonts}       
\usepackage{nicefrac}       
\usepackage{microtype}      
\usepackage{xcolor}         
\usepackage{dsfont}
\usepackage{amsmath}
\usepackage{graphicx}
\usepackage{makecell}
\usepackage{float}
\usepackage{amssymb}
\usepackage{subcaption}
\usepackage{multirow}

\newcommand{\method}{MetaWorld}
\newcommand{\eg}{e.g.,~}

\newcommand{\hut}{\textcolor{black}}
\title{MetaWorld: Scaling Multi-Agent Video World Model from Single-view Video Data}

\author{
Teng Hu$^1$\footnotemark[1]\thanks{Equal Contribution.}
\quad Mingchun Lu$^1$\footnotemark[1]
\quad Yating Wang$^1$
\quad Jiangning Zhang$^2$
\quad Jinkun Hao$^1$ \\
 \textbf{Ye Pan}$^1$
\quad \textbf{Ran Yi}$^1$\thanks{Corresponding Author.}
\quad \textbf{Lizhuang Ma}$^1$
\quad \textbf{Dacheng Tao}$^3$
\\
\normalsize $^1$Shanghai Jiao Tong University \quad $^2$Zhejiang University \quad $^3$Nanyang Technological University\\
{\tt\small Project page: \href{https://sjtuplayer.github.io/projects/MetaWorld/}{\textcolor{magenta}{https://sjtuplayer.github.io/projects/MetaWorld/}}}
}

\begin{document}

\maketitle

\vspace{-0.8cm}
\begin{figure}[H]
  \centering
  \includegraphics[width=1.0\textwidth]{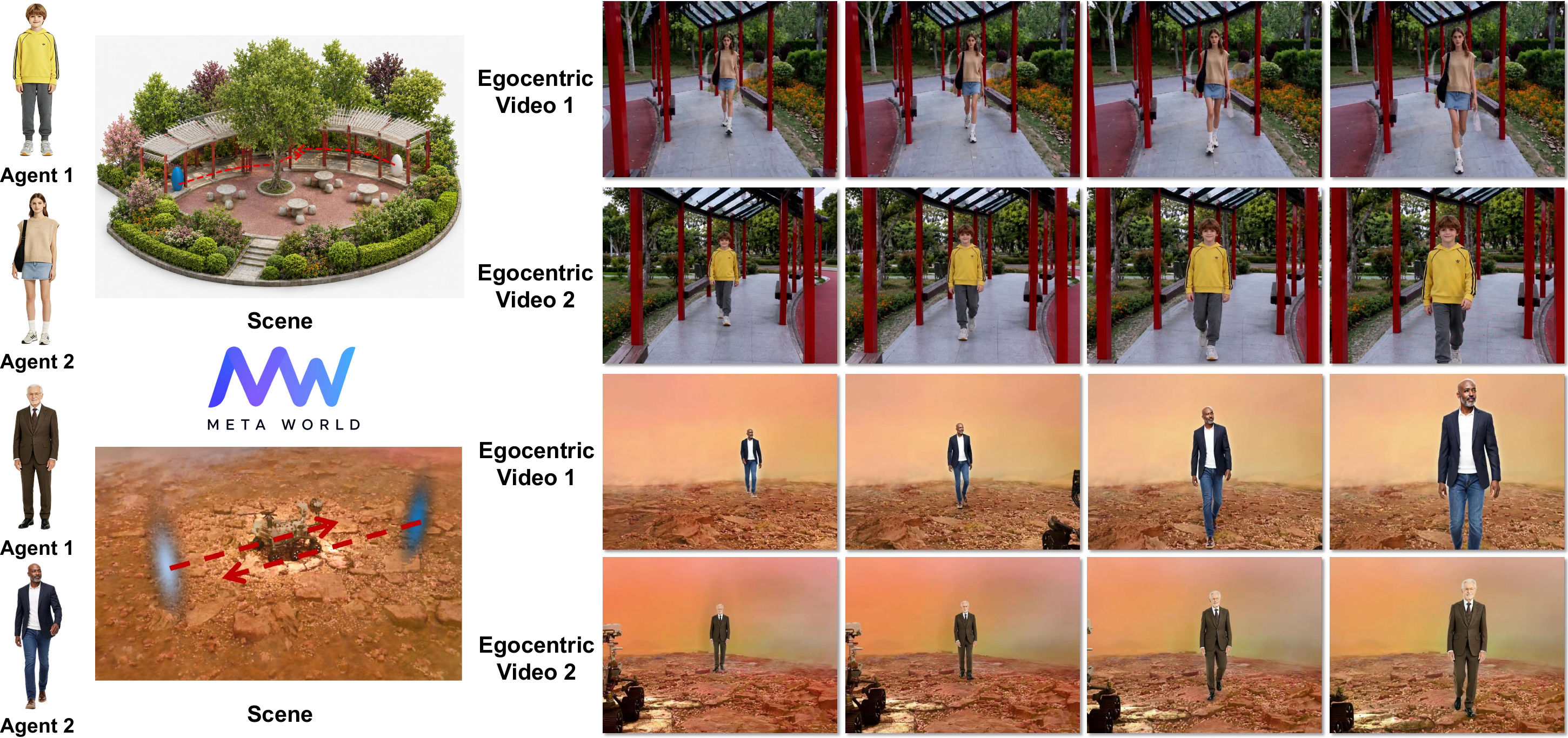}
  \vspace{-0.5cm}
\caption{\textbf{Multi-Agent Video World Modeling with \method.} 
Our framework successfully scales video world models to open-domain environments, generating identity-consistent video observations from multiple simultaneous egocentric perspectives. \method\ enforces both static geometric consistency and dynamic motion consistency, encouraging that the shared 3D physical environment evolves identically across all views. }
  \vspace{-0.2cm}
  \label{fig:teaser}
\end{figure}

\begin{abstract}
Video world models are a foundational generative technology for embodied AI and the Metaverse, yet existing approaches are inherently limited to a single agent observing from a single perspective. Extending these models to multi-agent settings introduces two critical challenges: \textbf{data scarcity} (coordinated multi-view recordings are prohibitively expensive to collect for general open-domain scenarios) and \textbf{world state alignment} (independently generated video streams cannot ensure that shared physical environments and events evolve consistently across views). To address these challenges, we propose \textbf{\method}, a novel framework that scales multi-agent video world models to open-domain environments directly from single-view videos. First, we introduce \textbf{Monocular World-State Unrolling (MWSU)} to explicitly decompose monocular footage into the camera operator's ego-motion and the visible subject's spatial trajectory. This camera-trajectory decomposition naturally extracts synchronized multi-agent motion data within a shared 3D space, completely bypassing the need for multi-camera setups. Second, for precise visual control, we develop the \textbf{Subject-Aware World Generator} to enable appearance-driven simulation conditioned on per-agent identity images. Finally, to ensure both views are grounded in the identical physical reality, we propose \textbf{World-State Alignment}, a per-frame inter-branch cross-attention mechanism inserted at every transformer layer of the video DiT. By jointly synchronizing the denoising process, WSA enforces both static geometric consistency and dynamic motion consistency, encouraging that the shared 3D environment and physical events remain well-aligned across both egocentric views. Extensive experiments demonstrate that \textbf{\method} achieves superior cross-view consistency and identity fidelity, establishing a highly scalable, physics-driven paradigm for multi-agent video world modeling.
\end{abstract}

\section{Introduction}
\label{sec:introduction}

Video generation has witnessed remarkable progress in recent years.
Driven by advances in diffusion models~\cite{ho2020denoising, rombach2022latent},
scalable Diffusion Transformers~\cite{peebles2023scalable, esser2024scaling},
and large-scale video pretraining~\cite{wan2025, kong2024hunyuanvideo,
xueultravideo,harmony,hu2025ultragen}, modern video generation systems can now synthesize
photorealistic, temporally coherent videos at unprecedented scale and quality.
This progress has in turn unlocked a new frontier: video world models~\cite{
lecun2022path}, systems that learn the dynamics of the physical world directly
from video data, enabling agents to predict future observations, reason about
consequences, and simulate complex environments without exhaustive real-world
interaction.
As a foundational generative technology, video world models have broad applications 
ranging from embodied AI~\cite{hafner2023mastering} to autonomous 
environment generation~\cite{parker2024genie2} and the Metaverse, which 
provides shared digital realities where multiple agents must interact within a 
synchronized, physics-driven space.

Existing video world models generally fall into two scene representation paradigms~\cite{survey2025}: \textbf{implicit 2D memory models}~\cite{parker2024genie2, sun2025worldplay} that generate future frames directly from learned representations, and \textbf{explicit 3D memory models}~\cite{survey2025} that rely on geometric structures and camera priors for spatial coherence. Crucially, both paradigms fundamentally assume a single agent observing the world from a single viewpoint. Extending these systems to \textbf{multi-agent video world modeling}, defined as the capability to simultaneously synthesize physically synchronized and identity-consistent egocentric video streams for multiple agents interacting within a shared environment, introduces two critical challenges: the extreme data scarcity of coordinated multi-agent egocentric recordings, and the strict requirement of world synchronization to ensure shared environments and dynamic events evolve identically across all views. While pioneering works like Solaris~\cite{solaris2025} and MultiWorld~\cite{multiworld} take initial steps toward multiplayer video world models, they are inherently confined to specific simulated environments (e.g., Minecraft). By relying on custom game engines to programmatically synchronize agents and collect multi-view observations, these approaches depend on privileged infrastructure that fails to generalize to open-domain settings. Overcoming this limitation requires a paradigm shift: scaling multi-agent dynamics directly from widely available single-view videos.


To build a generalizable multi-agent video world model for \textbf{open-domain environments} from single-view videos, we propose \textbf{\method}. Our framework directly addresses the aforementioned challenges through three core components.
First, to overcome data scarcity, \hut{we propose \textbf{Monocular World-State Unrolling (MWSU)} to extract hidden multi-agent physical dynamics directly from standard single-view videos. By explicitly decomposing this monocular footage into the unseen camera operator's ego-motion and the visible subject's spatial trajectory, we naturally acquire synchronized motion data of multiple agents navigating the exact same 3D space, thereby enabling scalable multi-agent training and inference within a shared scene via camera-trajectory decomposition.}
Second, to achieve precise visual control, we propose the \textbf{Subject-Aware World Generator (SAWG)}, a robust video diffusion backbone that conditions generation on per-agent identity images, enabling appearance-driven simulation where specific individuals' visual characteristics are preserved.
Finally, to ensure both views are grounded in the identical physical reality, we propose \textbf{World-State Alignment (WSA)}. WSA addresses the dual requirement of multi-agent simulation: \textbf{static geometric consistency} of the shared environment and \textbf{dynamic motion consistency} of evolving physical events. By coupling two parallel generation branches via a per-frame inter-branch cross-attention mechanism inserted at every transformer layer, WSA jointly synchronizes the denoising process. This encourages that both the 3D spatial layout and the temporal dynamics remain well-aligned across both agents' egocentric views.

\method\ has been evaluated on multi-camera egocentric benchmarks across
world consistency, mutual observability, and identity fidelity metrics.
Compared with single-agent baselines and ablated variants, \method\ achieves
significantly higher cross-view consistency while maintaining competitive
single-view generation quality, demonstrating the effectiveness of MWSU and WSA.
The contributions can be summarized as follows:

\begin{itemize}
  \item We propose \textbf{\method}, the first multi-agent video world model
    for \textbf{open-domain environments}. By successfully scaling from single-view
    videos, it addresses both the data scarcity and world state alignment challenges
    that have previously restricted multi-agent video world modeling to custom game engines.

    \item We introduce \textbf{Monocular World-State Unrolling (MWSU)}. By explicitly decomposing monocular footage into the camera operator's ego-motion and the subject's spatial trajectory, MWSU extracts synchronized multi-agent motion data within a shared 3D space, unlocking scalable multi-agent training via camera-trajectory decomposition.

  \item We propose the \textbf{Subject-Aware World Generator},
    a large pretrained video DiT fine-tuned to jointly support rendered condition
    video injection, trajectory conditioning, and identity image conditioning
    in a unified framework.

  \item We propose \textbf{World-State Alignment} to jointly synchronize 
    the dual-view generation process. By enforcing both static geometric consistency 
    and dynamic motion consistency, WSA ensures the shared 3D environment and physical 
    events remain well-aligned across both egocentric views.
\end{itemize}

\section{Related Work}
\label{sec:related_work}

\textbf{Video Generation Models.}
The emergence of large-scale Video Foundation Models (VFMs) has significantly revolutionized visual synthesis, driven by advancements in scalable Diffusion Transformers (DiTs)~\cite{peebles2023scalable, esser2024scaling} and massive text-video pre-training. Pioneering models such as Sora~\cite{sora}, CogVideoX~\cite{yang2024cogvideox}, HunyuanVideo~\cite{kong2024hunyuanvideo}, and Wan~\cite{wan2025} have demonstrated unprecedented capabilities in generating high-fidelity, spatiotemporally coherent videos. Alongside these advances, identity-conditioned customization has rapidly evolved to support precise single- and multi-subject synthesis~\cite{ConsisID, liu2025phantom, hu2025hunyuancustom, hu2025polyvivid}. Beyond mere visual synthesis, these powerful video models offer a robust foundation for building visual world simulators capable of understanding complex physical dynamics. Building upon them, \method\ realizes a multi-agent world model, seamlessly integrating precise identity control with physically consistent dual-view simulation.

\textbf{Video World Models.}
World models learn predictive models of environment dynamics to simulate future observations. Recent large-scale efforts, such as Genie~3~\cite{genie3} and Cosmos~\cite{cosmos}, have significantly advanced action-controllable and physically grounded simulations. Existing video world models generally fall into two scene representation paradigms. \textbf{Implicit memory world models}~\cite{huang2025selfforcing, li2025stable, cai2025mixture, yu2025context, xiao2025worldmem, sun2025worldplay} maintain historical consistency via streaming, autoregressive, or retrieval-based mechanisms. However, they lack explicit geometric constraints and are prone to structural inconsistency under complex camera movements. Conversely, \textbf{explicit 3D memory world models}~\cite{zhao2025spatia, li2025vmem, wu2025spmem, kong2025worldwarp, hong2025relic} integrate geometric structures (e.g., point clouds or 3D Gaussians) into the generation loop for stronger grounding, but typically require camera calibration, depth information, or known scene geometry. Crucially, these paradigms fundamentally assume a single observing agent. While recent pioneering works like Solaris~\cite{solaris2025} and Multiworld~\cite{multiworld} explore multi-agent world modeling, they are strictly confined to specific simulated environments (e.g., Minecraft) and heavily rely on synchronized multi-view data collected via programmable game engines. This strict data dependency severely limits their generalization to diverse, open-domain scenes. In contrast, \method\ introduces Dyadic Scene Decomposition (DSD) to build a highly generalizable multi-agent world model for \textbf{arbitrary real-world scenarios} directly from widely available single-view videos, bypassing the need for specialized multi-view data.

\section{Method}
\label{sec:method}

\begin{figure}[t]
  \centering
  \includegraphics[width=1.0\textwidth]{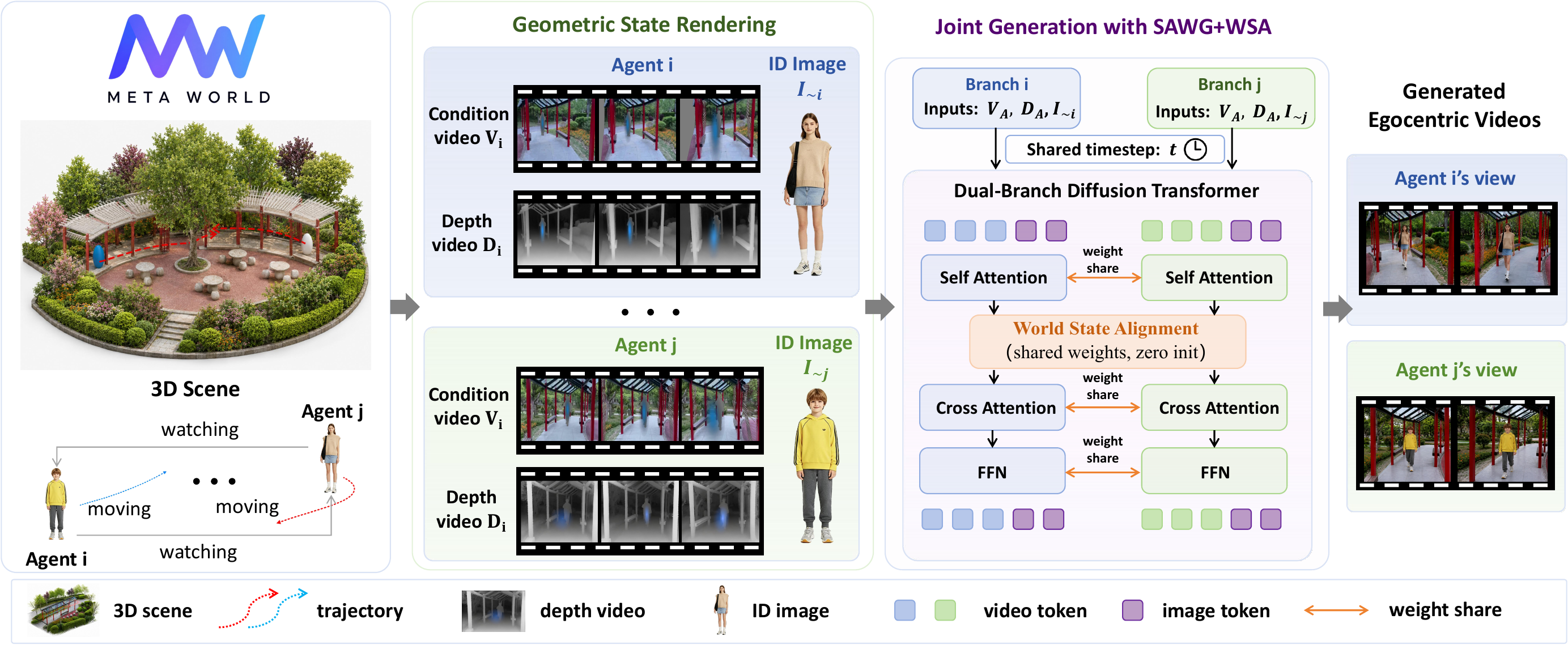}
  \vspace{-0.15in}
\caption{\textbf{Framework of \method.} The framework takes agent-specific geometric priors (RGB condition videos and depth videos) rendered from the shared 3D world, alongside corresponding identity (ID) reference images, as inputs. These conditions are processed by parallel branches of our Subject-Aware World Generator. Synchronized by the World-State Alignment (WSA) module during joint denoising, the model generates physically consistent and identity-preserving egocentric videos as observed by multiple agents simultaneously.}
  \vspace{-0.15in}
  \label{fig:framework}
\end{figure}


\method\ addresses the task of \textbf{multi-agent video world modeling} for open-domain environments. 
Given a shared scene containing $N$ agents, the system takes their absolute global trajectories as input. 
For each agent $i$, we compute the relative trajectories of the other agents projected onto agent $i$'s camera perspective, rendering them into a condition video. 
The model takes these $N$ condition videos, alongside the identity (ID) reference images of the visible agents, and jointly outputs $N$ egocentric video observations. 
The goal is to ensure that these generated views are individually photorealistic while remaining strictly consistent with respect to the shared physical world.
The core innovation of \method\ lies in its ability to scale this multi-agent simulation directly from widely available \textbf{single-view videos}. 
The overall framework is illustrated in Fig.~\ref{fig:framework}.

\subsection{Multi-Agent World Generation via Monocular World-State Unrolling}
\label{subsec:unrolling}

\textbf{The multi-agent data bottleneck.}
Building a multi-agent video world model requires data where multiple agents simultaneously capture egocentric views of the same scene. However, synchronized multi-agent recordings are prohibitively expensive to collect and scale across diverse open-domain environments. In contrast, single-view egocentric videos are naturally abundant in datasets, social media, and sports footage. The critical challenge lies in bridging the gap between this abundant monocular data and the strict multi-view supervision required for training a multi-agent system.

\textbf{Monocular World-State Unrolling (MWSU).}
\hut{To address this bottleneck, MWSU leverages the insight that a single-view egocentric video capturing another active person already inherently encodes the physical dynamics of a multi-agent interaction. It simultaneously contains the continuous spatial state of the unseen camera operator (the observer) and the visible subject (the observed). MWSU explicitly extracts this hidden multi-agent world state directly from monocular footage. By \textbf{decomposing} the video into independent physical parameters, specifically extracting the camera's ego-motion to represent one agent and parameterizing the spatial trajectory to represent the other, we successfully decouple the observer from the observed. This physical decomposition allows us to systematically \textbf{recombine} these isolated elements to construct rigorous multi-agent interactions, obtaining the exact coordinated multi-agent states needed to supervise our multi-agent world model. Concretely, for a \textbf{shared scene} $\mathcal{S}_i$ with $N$ agents, any egocentric video $\mathbf{x}_i$ recorded by agent $i$ is decomposed into three structural components:}
\begin{itemize}
  \item \textbf{Camera motion sequence}
    $\mathbf{c}_i \in \mathbb{R}^{T \times D}$: the per-frame camera poses defining agent $i$'s observation over $T$ timesteps, serving as the primary spatial condition.
  \item \textbf{Multi-subject trajectory set}
    $\mathcal{T}_{\sim i}^{(i)} = \{ \boldsymbol{\tau}_j^{(i)} \}_{j \neq i}$: the $D'$-dimensional per-frame positions $\{P_j\}$ of all other visible agents $j$, \emph{projected onto} agent $i$'s camera space $\mathbf{c}_i$. This defines how subjects move within $i$'s field of view.
  \item \textbf{Identity reference set} 
    $\mathcal{I}_{\sim i} = \{ \mathbf{I}_j \}_{j \neq i}$: reference images encoding the appearance of each visible agent $j$ to ensure visual consistency throughout generation.
\end{itemize}
Formally, agent $i$'s egocentric video is modeled as a conditional distribution:
\begin{equation}
  \mathbf{x}_i \;\sim\; p_\theta\,\!\bigl(\mathbf{x}_i
    \;\big|\; \mathcal{S}_i,\; \mathbf{c}_i,\; \mathcal{T}_{\sim i}^{(i)},\; \mathcal{I}_{\sim i}\bigr).
  \label{eq:unrolling_i}
\end{equation}

\textbf{Multi-agent inference.}
\hut{At inference time, our goal is to generate a synchronized $N$-agent simulation $\{ \mathbf{x}_1, \dots, \mathbf{x}_N \}$ from a unified global scene. This shared scene is fully specified by a set of $N$ global agent trajectories $\{ \mathbf{P}_1, \dots, \mathbf{P}_N \}$ and their corresponding identity reference images $\{ \mathbf{I}_1, \dots, \mathbf{I}_N \}$. To leverage the learned conditional distribution $p_\theta$ defined in Eq.~\eqref{eq:unrolling_i}, we must deterministically construct the exact condition tuple $(\mathcal{S}_i, \mathbf{c}_i, \mathcal{T}_{\sim i}^{(i)}, \mathcal{I}_{\sim i})$ for every agent $i$. }
Crucially, an agent's physical movement in the 3D scene dually dictates its own egocentric view and its visible motion to others. Thus, for any given agent $i$, its global trajectory $\mathbf{P}_i$ directly translates into its camera motion sequence $\mathbf{c}_i$. Subsequently, all cross-view conditions are geometrically determined: projecting any other agent $j$'s global trajectory $\mathbf{P}_j$ onto $\mathbf{c}_i$ directly yields the relative trajectory $\boldsymbol{\tau}_j^{(i)}$, successfully formulating the trajectory set $\mathcal{T}_{\sim i}^{(i)}$. Together with the rendered 3D background $\mathcal{S}_i$ and the gathered identity set $\mathcal{I}_{\sim i}$, the full conditioning requirement for Eq.~\eqref{eq:unrolling_i} is satisfied. 
The single trained model $p_\theta$ is then applied symmetrically and simultaneously across all $N$ branches. This joint sampling process effectively translates a unified set of physical world parameters into a coherent, multi-view simulation.

\textbf{Scalable MWSU Data Engine.}
We propose an automated, geometry-aware data engine to instantiate the MWSU training tuple $(\mathcal{S}_i,\mathbf{c}_i, \mathcal{T}_{\sim i}^{(i)}, \mathcal{I}_{\sim i}, \mathbf{x}_i)$ from raw single-view video $\mathbf{x}_i$. 
First, we extract the camera motion sequence $\mathbf{c}_i$ using MoGe-2~\cite{wang2025moge}. 
Simultaneously, we track the set of dynamic subjects within the video using SAM~2~\cite{ravi2024sam} to obtain continuous instance masks, and estimate per-frame depth using Depth Anything~\cite{yang2024depth}. 
From these tracking results, we select a single frame with a valid, non-empty mask for each individual subject $j$ to explicitly crop out their visual appearance. This yields the individual identity reference image $\mathbf{I}_j$, which collectively form the identity set $\mathcal{I}_{\sim i}$.
By combining the continuous masks, depth maps, and camera poses $\mathbf{c}_i$, we calculate the global 3D trajectory coordinates for each visible agent. 
These global trajectories are then projected back onto the camera operator's perspective to formulate the relative trajectory set $\mathcal{T}_{\sim i}^{(i)}$, where each subject's position is visually represented as a moving 3D Gaussian sphere.
To provide environmental context, we construct the shared 3D background scene $\mathcal{S}_{i}$ using two distinct strategies to supervise different model capabilities (\textit{details are in the appendix}): 
(1) \textbf{Anchor-Frame Background Modeling}: utilizing only the initial frame to establish a partial environmental prior, which explicitly trains the model's capacity for scene completion and spatial extrapolation; 
(2) \textbf{Holistic Background Aggregation}: accumulating static background geometries across all frames (with dynamic objects masked out) to build a comprehensive 3D prior, which can enhance the spatial consistency over time. 

\subsection{Subject-Aware World Generator}
\label{subsec:base_model}


To learn the universal MWSU distribution $p_\theta\,\!\bigl(\mathbf{x}_i \;\big|\; \mathcal{S}_i,\; \mathbf{c}_i,\; \mathcal{T}_{\sim i}^{(i)},\; \mathcal{I}_{\sim i}\bigr)$ derived from the data engine, we introduce the \textbf{Subject-Aware World Generator (SAWG)}. Built upon the powerful Phantom-Wan2.1 14B DiT backbone~\cite{liu2025phantom, wan2025}, \hut{SAWG transcends standard video generation by acting as a highly controllable, physics-grounded simulator. Rather than merely concatenating conditional inputs, SAWG is explicitly designed to disentangle and independently govern \emph{where} a subject moves and \emph{who} the subject is. To achieve this, our conditioning pipeline operates through three synergistic stages: first, \textbf{Geometric State Rendering} translates the abstract 3D environment and multi-agent trajectories into dense RGB and depth priors; second, \textbf{Subject-Location and Geometry Grounding} injects these spatial conditions into the latent space to enforce physical and kinematic boundaries; finally, \textbf{Subject-Appearance Anchoring} integrates persistent identity references to improve zero-shot visual fidelity. This decoupled yet unified design ensures that the generated subjects adhere flawlessly to designated 3D trajectories while maintaining consistent visual characteristics throughout complex dynamic environments.}

\textbf{Geometric State Rendering.}
Following the data engine pipeline, we construct the specific visual conditions for each agent. First, we sample the constructed 3D background along the camera motion sequence $\mathbf{c}_i$ to generate a base scene video. We then explicitly superimpose the moving 3D Gaussian spheres, which represent the relative trajectory set $\mathcal{T}_{\sim i}^{(i)}$, into this visual space. This combined process yields the RGB condition RGB video $V \in \mathbb{R}^{C \times T \times H \times W}$. Simultaneously, we render the corresponding depth video $D \in \mathbb{R}^{1 \times T \times H \times W}$, which captures the exact per-frame depth maps of both the static background and the superimposed Gaussian trajectory spheres. Together with the identity reference set $\mathcal{I}_{\sim i}$, these rendered streams form the complete conditioning input for the generation of the target egocentric video $\mathbf{x}_i$.

\textbf{Subject-Location and Geometry Grounding.}
To explicitly inject both the established physical environment and the precise spatial locations of the subjects into the generative process, the RGB condition video $V$ and the depth video $D$ are compressed by the VAE encoder $\mathcal{E}$ and concatenated along the channel dimension. This forms a joint geometric representation that explicitly encodes the subjects' moving coordinates within the 3D layout:
\begin{equation}
  \mathbf{z}_{\mathrm{cond}} = \bigl[\mathcal{E}(V);\;\mathcal{E}(D)\bigr]
    \;\in\; \mathbb{R}^{2C_z \times T_z \times H_z \times W_z}.
  \label{eq:cond_cat}
\end{equation}
To match the latent space of the diffusion model, this joint representation is processed by a zero-initialized feature compression network $\mathcal{F}_{\mathrm{zero}}$, which projects the channels from $2C_z$ back to $C_z$. The compressed geometric features are then residually added to the noisy video latent $\mathbf{z}_t$:
\begin{equation}
  \mathbf{z}'_t = \mathbf{z}_t + \mathcal{F}_{\mathrm{zero}}(\mathbf{z}_{\mathrm{cond}}).
  \label{eq:traj_inject}
\end{equation}
By employing zero-initialization, we ensure that the Phantom-Wan2.1 backbone fully retains its powerful pretrained generative priors at the onset of fine-tuning. This design choice allows the structural scene context and the kinematic subject trajectory guidance to emerge gradually, ensuring highly stable training dynamics.

\textbf{Subject-Appearance Anchoring.}
While the geometric grounding dictates \emph{where} the subjects are, SAWG must also guarantee \emph{who} they are. To achieve the promised zero-shot identity fidelity, the identity reference images from $\mathcal{I}_{\sim i}$ must serve as unwavering visual conditions. We encode these images via the VAE and concatenate them directly to the geometry-grounded video latent $\mathbf{z}'_t$ along the temporal axis. For an individual subject $j$, this injection is formulated as:
\begin{equation}
  \tilde{\mathbf{z}}_t = \bigl[\mathbf{z}'_t;\;\mathcal{E}(\mathbf{I}_j)\bigr]
    \;\in\; \mathbb{R}^{C_z \times (T_z+1) \times H_z \times W_z}.
  \label{eq:id_inject}
\end{equation}
This formulation extends naturally to $N$-agent scenarios by simply appending additional latent identity frames. Crucially, this mechanism acts as a persistent appearance anchor, supplying the DiT with continuous, frame-independent visual guidance for each specific subject at every denoising step, while the training loss is backpropagated exclusively over the $T_z$ target video frames.


\subsection{World-State Alignment}
\label{subsec:wsa}

\textbf{The multi-view inconsistency problem.}
Running $N$ independent SAWG models (one for each agent's view) leads to two critical failures in multi-agent simulation. The first is \textbf{geometric inconsistency}, where the static 3D spatial layout (e.g., background structures, stationary objects) diverges across views. The second is \textbf{dynamic inconsistency}, where shared transient events (e.g., moving objects, lighting changes, or physical interactions) fail to synchronize. Because trajectory conditioning only dictates individual subject motion, it cannot resolve these shared environmental discrepancies. To achieve coherent multi-agent simulation, the parallel generation branches must be fundamentally grounded in an identical, synchronized physical world state.

\textbf{Frame-wise inter-branch cross-attention.}
To explicitly synchronize this shared world state across different branches, we propose \textbf{World-State Alignment (WSA)}. Building upon the Wan2.1 backbone, we introduce a dedicated \textbf{frame-wise inter-branch cross-attention} module into every transformer layer. By forcing parallel generation branches to continuously exchange latent representations, this mechanism ensures that both coarse spatial layouts in early layers and fine dynamic details in deeper layers remain well-aligned. 
Taking target branch $i$ and source branch $j$ as an example, when branch $i$ queries world-state information from branch $j$, the WSA computation is formulated as:
\begin{equation}
  \mathbf{x}_i \leftarrow \mathbf{x}_i
    + \mathrm{WSAAttn}\!\bigl(\mathrm{LN}_{\mathrm{d}}(\mathbf{x}_i),\;
       \mathrm{KV}{=}\mathrm{LN}_{\mathrm{d}}(\mathbf{x}_j)\bigr).
  \label{eq:wsa_attn}
\end{equation}
To prevent cross-frame temporal leakage and preserve physical causality, WSA strictly restricts attention to \textbf{within-frame token pairs}: the $H_z W_z$ tokens of frame $f$ in branch $i$ attend exclusively to those of frame $f$ in branch $j$. This leaves only 2D spatial relative positions as effective positional signals, allowing WSA to learn cross-view spatial correspondences effectively.

\begin{figure}[t]
  \centering
  \includegraphics[width=1.0\textwidth]{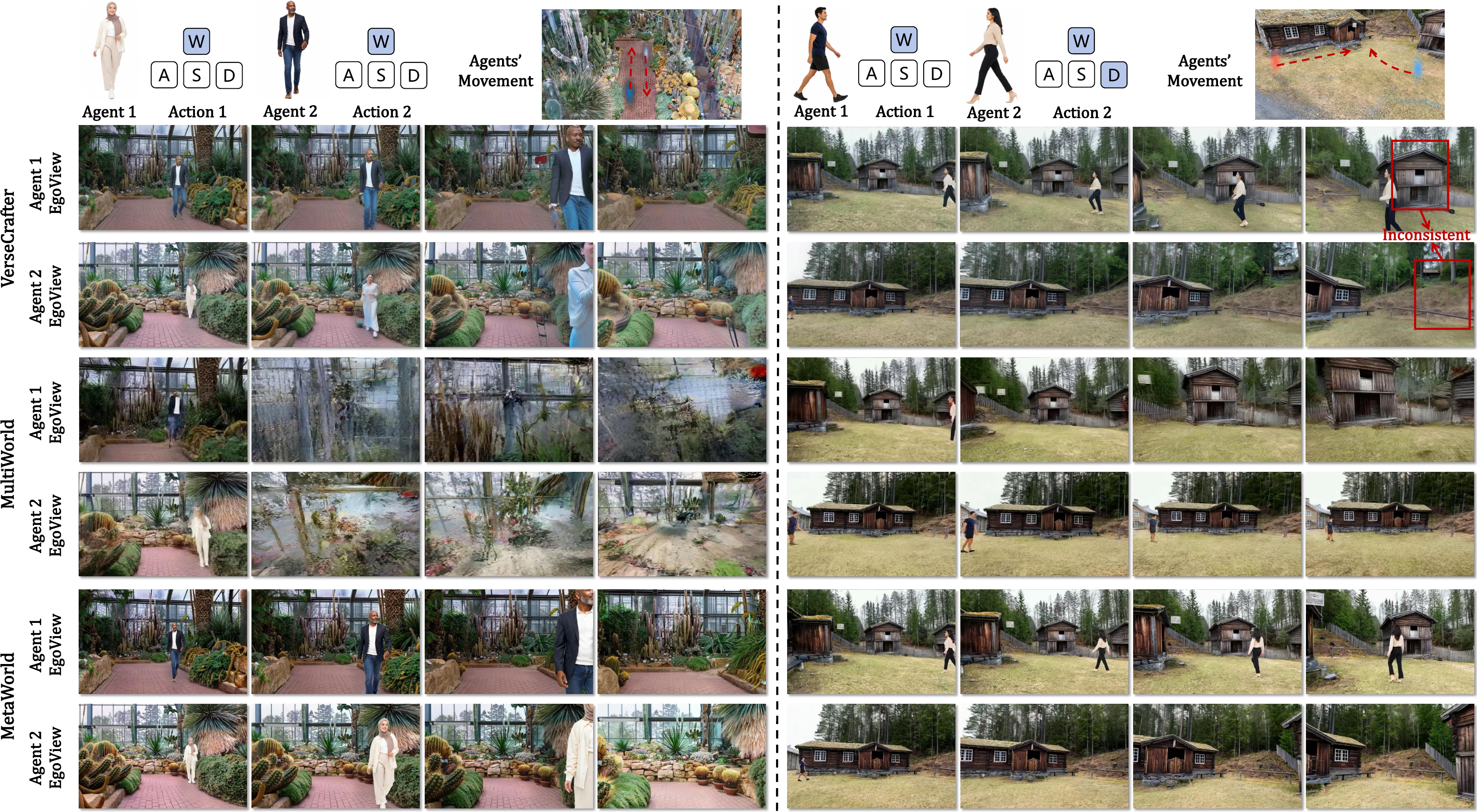}
  \vspace{-0.1in}
  \caption{Qualitative comparison on multi-view generation with the state-of-the-arts.}
  \vspace{-0.15in}
  \label{fig:compare}
\end{figure}

\textbf{Dual-agent training strategy.}
While our ultimate objective is $N$-agent simulation, we efficiently train the WSA module using paired dual-agent interactions. To achieve this, we collect a small-scale dataset of real-world multi-view videos and process them through our previously described Scalable MWSU Data Engine. This extracts the precise multi-agent training tuples (e.g., aligned camera poses, relative trajectories, and identities) required for synchronized supervision. During training, we instantiate two branches ($A$ and $B$) and compute the cross-attention bidirectionally ($A$ attends to $B$, and $B$ attends to $A$). To preserve the base model's pretrained generative priors, the output projection of the WSA module is zero-initialized.
Both branches share the same denoising timestep $t$, and the training loss is computed as the average of their individual flow-matching losses: $\mathcal{L}_{\mathrm{WSA}} = \frac{1}{2}\bigl(\mathcal{L}_A + \mathcal{L}_B\bigr)$.

\textbf{Multi-agent cyclic inference.}
At inference time, we elegantly scale the dual-trained WSA to $N$ agents by adopting a \textbf{directed cyclic attention topology}. Instead of computing computationally prohibitive $O(N^2)$ dense cross-attention across all agents, we sequentially inject physical world-state information. Specifically, agent 1's latent state is injected as the KV condition for agent 2, agent 2 into agent 3, and this chain continues until agent $N$ injects its state back into agent 1. Formally, for any agent $k \in \{1, \dots, N\}$, the cross-attention is computed as:
\begin{equation}
  \mathbf{x}_k \leftarrow \mathbf{x}_k
    + \mathrm{WSAAttn}\!\bigl(\mathbf{x}_k,\; \mathrm{KV}{=}\mathbf{x}_{(k-2 \bmod N) + 1}\bigr).
  \label{eq:cyclic_infer}
\end{equation}
This cyclic topology efficiently propagates global consistency across all views. By running parallel sampling with tightly synchronized timesteps, the model yields mutually coherent, physics-grounded multi-agent video streams from a single unified simulation.

\section{Experiments}
\label{sec:experiment}

\subsection{Experiment Settings}

\textbf{Implementation details.}
\method\ is built on the Phantom-Wan2.1 14B DiT backbone~\cite{liu2025phantom,wan2025}.
In Stage~1, we train the subject-aware world generator on single-view data
using AdamW with a constant learning
rate of $5{\times}10^{-4}$,
and \texttt{bf16} mixed precision on 32$\times$ H20 GPUs.
In Stage~2, we initialize from the Stage-1 checkpoint and introduce the
World-State Alignment module for synchronization, training with the same
optimizer and learning rate on 32$\times$ H20 GPUs.
Both stages generate 81-frame videos at 480P resolution.
We use flow matching with logit-normal weighting (flow shift $\gamma{=}5.0$)
and perform inference with 40-step Euler sampling at a CFG scale of $\eta{=}5.0$.


\textbf{Baselines.}
Since no prior multi-agent video world models generalize to open-domain scenarios, we adapt two representative methods for comparison:
\textbf{(1) MultiWorld}~\cite{multiworld}: A multi-view video generation model. We directly utilize its official pretrained weights to evaluate its performance under the testset.
\textbf{(2) VerseCrafter}~\cite{versecrafter}: An image-to-video model supporting fine-grained camera and trajectory controls. We simulate multi-agent generation by conditioning it on each agent's respective trajectories to render views independently. For a strict and fair comparison, VerseCrafter is initialized with the exact first frame generated by our model.

\textbf{Metrics.}
We adopt the VBench~\cite{huang2024vbench} suite to assess general video generation quality, specifically reporting \textbf{Subject Consistency}, \textbf{Background Consistency}, \textbf{Motion Smoothness}, \textbf{Aesthetic Quality}, and \textbf{Imaging Quality}. Detailed definitions for these standard metrics are provided in the supplementary material. Furthermore, to evaluate multi-agent synchronization and spatial control, we introduce two specific metrics. \textbf{Cross-View Consistency (CVC)} measures the DINO-v2 feature similarity between corresponding regions across the generated views, explicitly evaluating whether the cameras observe physically consistent scene content. Finally, \textbf{Trajectory Consistency} assesses the accuracy of our motion conditioning by calculating the alignment between the input spatial trajectories and the actual physical movements extracted from the generated videos, ensuring the agents strictly adhere to their specified pathways.

\subsection{Comparison Results}


\textbf{Qualitative comparison.}
Figure~\ref{fig:compare} presents a visual comparison of the generated multi-view sequences. While VerseCrafter~\cite{versecrafter} utilizes strong image-to-video priors, it struggles to maintain strict identity consistency, frequently generating videos where the subject's appearance diverges from the specified agent ID. Furthermore, as illustrated in the rightmost case, it fails to preserve cross-view environmental consistency, resulting in the two agents observing  contradictory background scenes. MultiWorld~\cite{multiworld}, on the other hand, exhibits a limited capability in modeling complex open-domain environments; its generated videos suffer from noticeable artifacts and an overall lower visual quality. In contrast, \method\ successfully generates multi-view videos that maintain strict identity fidelity while ensuring flawless structural and dynamic synchronization across all camera branches.

\textbf{Quantitative comparison.}
We quantitatively evaluate all methods under two distinct settings: Full-3D scene and Partial-scene, as reported in Tab.~\ref{tab:comparison}. 
In the \textbf{Full-3D scene setting}, the complete 3D spatial layout is provided. While VerseCrafter achieves good single-view aesthetic quality, it struggles with cross-view geometry consistency due to its lack of explicit multi-agent world synchronization. MultiWorld attempts to align the multi-agent views but still falls short of \method, whose WSA mechanism explicitly couples the denoising process to guarantee an identical shared reality.
The \textbf{Partial-scene setting} reflects highly realistic scenarios where comprehensive 3D assets are unavailable, and the models must dynamically hallucinate unobserved content based on a partial environment reconstructed from just a single image or sparse views. In this setting, VerseCrafter generates visually plausible extrapolations but struggles to coordinate these hallucinated regions globally, occasionally producing contradictory out-of-view content between the cameras. MultiWorld shows structural alignment on explicitly provided regions but struggles to accurately synchronize the hallucinated portions. \method\ achieves the highest CVC across both settings, demonstrating that our synchronized joint denoising enables the parallel branches to consistently agree on both the specified physical geometry and the unspecified hallucinated content.

\begin{table}[t]
  \centering
  \caption{
    Quantitative comparison across two evaluation settings.
    \textbf{Bold} and \underline{underline} denote the best and second-best results, respectively.}
  \resizebox{0.95\linewidth}{!}{
    \begin{tabular}{llccccccc}
      \toprule
      Setting & Method
        & \makecell{Subject\\Consist.$\uparrow$}
        & \makecell{Background\\Consist.$\uparrow$}
        & \makecell{Motion\\Smooth.$\uparrow$}
        & \makecell{Aesthetic\\Quality$\uparrow$}
        & \makecell{Imaging\\Quality$\uparrow$}
        & CVC$\uparrow$ 
        & \makecell{Trajectory\\Consist.$\uparrow$} \\
      \midrule
      \multirow{3}{*}{Full-3D}
        & VerseCrafter~\cite{versecrafter} & \underline{0.9299} & \underline{0.9071} & \textbf{0.9891} & \underline{0.4999} & \underline{0.7015} & \underline{0.7210} & \underline{0.9634} \\
        & MultiWorld~\cite{multiworld}     & 0.7999 & 0.8945 & 0.9820 & 0.4417 & 0.5998 & 0.3631 & 0.2355 \\
        & \textbf{\method\ (Ours)}         & \textbf{0.9335} & \textbf{0.9171} & \underline{0.9854} & \textbf{0.5817} & \textbf{0.7574} & \textbf{0.8454} & \textbf{0.9752} \\
      \midrule
      \multirow{3}{*}{Partial-3D}
        & VerseCrafter                     & \underline{0.9229} & \underline{0.9015} & \textbf{0.9865} & \underline{0.4905} & \underline{0.6973} & \underline{0.7280} & \underline{0.9582} \\
        & MultiWorld                       & 0.7915 & 0.8907 & 0.9778 & 0.4349 & 0.5902 & 0.4081 & 0.2265 \\
        & \textbf{\method}                 & \textbf{0.9289} & \textbf{0.9107} & \underline{0.9818} & \textbf{0.5735} & \textbf{0.7526} & \textbf{0.7549} & \textbf{0.9694} \\
      \bottomrule
    \end{tabular}
  }
  \label{tab:comparison}
  \vspace{-0.15in}
\end{table}

\subsection{Ablation Study}

To validate the core architectural designs and data processing strategies of our framework, we conduct a comprehensive ablation study. Quantitative results are provided in Tab.~\ref{tab:ablation}.

\textbf{Effect of World-State Alignment (w/o WSA).}
We ablate the WSA module, forcing the model to generate each agent's egocentric view entirely independently. While the individual trajectory conditioning manages to guide basic movement and single-view generation qualities remain highly competitive (even achieving high Background Consistency and Imaging Quality), the lack of frame-wise inter-branch cross-attention completely severs the communication of shared background dynamics. Consequently, the model fails to synchronize transient environmental events, resulting in a distinct drop in Cross-View Consistency (CVC).

\textbf{Importance of depth priors (w/o Depth Video).}
We evaluate the necessity of explicit geometric grounding by removing the depth video $\mathbf{d}$ from the condition injection. Without depth priors, the model lacks the structural references needed to accurately anchor the projected 3D spatial trajectories within the scene's layout. This ambiguity leads to severe floating artifacts and a big drop in \textbf{Trajectory Consistency}, as the generated agents fail to strictly adhere to the specified physical pathways.

\textbf{Impact of background modeling strategies.}
Our MWSU data engine relies on two complementary strategies for spatial conditioning. We ablate them individually:
(1) \textbf{w/o Anchor-Frame BG}: Removing the initial anchor-frame prior strips the model of explicit supervision for scene extrapolation. Consequently, the model struggles to hallucinate and fill in unobserved regions naturally, leading to a noticeable degradation in overall generation quality (particularly in Aesthetic and Imaging Quality).
(2) \textbf{w/o Holistic BG}: Relying only on the initial frame without accumulating the holistic 3D background geometry causes the model to lose global spatial awareness. While single-view generation metrics remain competitive and close to the state-of-the-art, the model utterly fails to follow the designated scene layout under complex trajectories with large camera rotations, resulting in a severe drop in CVC.

\textbf{Impact of zero-initialization (w/o ZeroInit).}
We replace the zero-initialization strategy in both the WSA output projection and the feature compression network with standard random initialization. This abrupt alteration of the latent space severely disrupts the powerful pretrained generative priors of the Wan2.1 backbone. Rather than diverging completely, the training struggles with instability and ultimately converges to a suboptimal local minimum, resulting in a roughly 20\% performance degradation across all evaluated metrics.
\begin{table}[t]
  \centering
  \caption{Ablation study on key design choices of \method. 
    \textbf{Bold} denotes the best result.}
  \renewcommand{\arraystretch}{1.1}
  \resizebox{0.9\linewidth}{!}{
    \begin{tabular}{l|ccccccc}
      \toprule
      Variant 
        & \makecell{Subject\\Consist.$\uparrow$}
        & \makecell{Background\\Consist.$\uparrow$}
        & \makecell{Motion\\Smooth.$\uparrow$}
        & \makecell{Aesthetic\\Quality$\uparrow$}
        & \makecell{Imaging\\Quality$\uparrow$}
        & CVC$\uparrow$ 
        & \makecell{Trajectory\\Consist.$\uparrow$} \\
      \midrule
      w/o WSA                  & \underline{0.9310} & \textbf{0.9148} & \underline{0.9826} & 0.5636 & \textbf{0.7620} & 0.7358 & \underline{0.9542} \\
      w/o Depth Video          & 0.9282 & 0.9110 & 0.9812 & \underline{0.5688} & 0.7445 & \underline{0.7840} & 0.9126 \\
      w/o Anchor-Frame BG      & 0.9075 & 0.8455 & 0.9826 & 0.4588 & 0.6812 & 0.7496 & 0.9284 \\
      w/o Holistic BG          & 0.9218 & 0.8821 & 0.9806 & 0.5621 & 0.6170 & 0.5513 & 0.9352 \\
      w/o ZeroInit             & 0.7583 & 0.7630 & 0.9568 & 0.3820 & 0.6358 & 0.6172 & 0.7845 \\
      \midrule
      \textbf{\method\ (Ours)} & \textbf{0.9312} & \underline{0.9139} & \textbf{0.9836} & \textbf{0.5776} & \underline{0.7550} & \textbf{0.8001} & \textbf{0.9723} \\
      \bottomrule
    \end{tabular}
  }
  \label{tab:ablation}
  \vspace{-0.15in}
\end{table}

\section{Conclusion}
\label{sec:conclusion}

In this paper, we present \method, the first multi-agent video world model capable of simulating open-domain environments. To overcome the severe data scarcity that has traditionally confined multi-agent world modeling to custom game engines like Solaris~\cite{solaris2025}, we introduce \textbf{Monocular World-State Unrolling (MWSU)}. Grounded in the insight that standard single-view egocentric videos implicitly encode complete physical dual-agent dynamics, MWSU directly extracts camera ego-motion and subject trajectories. This fundamentally shifts the paradigm by allowing scalable multi-agent supervision without the need for coordinated multi-camera recordings. 
Building upon this data engine, our \textbf{Subject-Aware World Generator (SAWG)} functions as a robust video diffusion backbone, enabling precise appearance-driven simulation that strictly preserves individual visual characteristics. Finally, to ensure all simulated views are grounded in an identical physical reality, we propose \textbf{World-State Alignment (WSA)}. By coupling parallel generation branches via a frame-wise inter-branch cross-attention mechanism, WSA successfully enforces both the static geometric consistency of the shared 3D environment and the dynamic motion consistency of evolving events. Extensive experiments demonstrate that \method\ significantly outperforms prior baselines in cross-view consistency, physical synchronization, and identity fidelity. Our framework advances the state of the art for world models, opening new possibilities for collaborative embodied AI, synchronized Metaverse environments, and scalable multi-agent video generation.

\bibliography{references}
\bibliographystyle{abbrvnat}




\clearpage

\appendix
\section{Overview}
\label{sec:suppl_overview}

In the appendix, we offer further details on implementation, present additional experimental results, and provide more comprehensive analyses, structured as follows:
\begin{itemize}
  \item Implementation details (Sec.~\ref{sec:suppl_impl});
  \item Dataset Construction (Sec.~\ref{sec:dataset construction})
  \item Single-world generation results (Sec.~\ref{sec:suppl_single});
  \item Additional multi-agent generation results (Sec.~\ref{sec:suppl_more_compare});
  \item Limitations (Sec.~\ref{sec:suppl_limitations}).
\end{itemize}

\section{Implementation Details}
\label{sec:suppl_impl}

\subsection{Hyperparameters.}
Table~\ref{tab:hparams} summarizes the key hyperparameters utilized during both the training (Stage 1 and Stage 2) and inference phases of our framework.
\begin{table}[h]
  \centering
  \caption{Training hyperparameters for Stage 1 and Stage 2.}
  \label{tab:hparams}
  \begin{tabular}{lll}
    \toprule
    Hyperparameter & Stage 1 & Stage 2 \\
    \midrule
    Backbone init       & Phantom-Wan2.1 14B       & Stage-1 checkpoint \\
    New params          & $\mathcal{F}_{\mathrm{zero}}$ (zero-init conv)
                        & WSA cross-attn layers \\
    Optimizer           & AdamW                    & AdamW \\
    $(\beta_1, \beta_2)$ & $(0.9, 0.999)$          & $(0.9, 0.999)$ \\
    Learning rate       & $5{\times}10^{-4}$       & $5{\times}10^{-4}$ \\
    LR schedule         & Constant                 & Constant \\
    Mixed precision     & \texttt{bf16}            & \texttt{bf16} \\
    Distributed         & DeepSpeed ZeRO           & DeepSpeed ZeRO \\
    Video length        & 81 frames                & 81 frames \\
    Resolution          & $480{\times}832$         & $480{\times}832$ \\
    $C_z$               & 16                       & 16 \\
    Flow shift $\gamma$ & 5.0                      & 5.0 \\
    CFG scale $\eta$    & 5.0                      & 5.0 \\
    Inference steps     & 40                       & 40 \\
    GPU                 & 32$\times$ H20     & 32$\times$ H20 \\
    \bottomrule
  \end{tabular}
\end{table}

\subsection{Metrics}

\textbf{Visual Quality.}
We adopt the VBench~\cite{huang2024vbench} evaluation suite to assess generated video quality from multiple dimensions. Specifically, we report:
\textbf{Subject Consistency}, which measures the identity coherence of subjects across frames using DINO features;
\textbf{Background Consistency}, which evaluates the temporal stability of background regions via CLIP features;
\textbf{Motion Smoothness}, which quantifies the smoothness of inter-frame motion;
\textbf{Aesthetic Quality}, which scores the visual appeal of generated frames;
and \textbf{Imaging Quality}, which captures low-level image quality (\eg sharpness, noise level).
In addition, we introduce a \textbf{Cross-View Consistency (CVC)} metric to evaluate whether the generated views preserve a consistent underlying 3D scene structure and appearance.

\textbf{Cross-View Consistency.}
While VBench evaluates single-video quality, it does not directly measure whether two generated views describe the same underlying 3D scene. We therefore introduce \textbf{Cross-View Consistency (CVC)}, a point-cloud-based metric that evaluates both geometric and appearance consistency across views. Given a reference video $V_a$ and a generated video $V_b$, we reconstruct a colored point cloud from each video using monocular depth estimation~\cite{yang2024depth} and camera back-projection. Both videos are truncated to the first 3 seconds for evaluation.

For each frame $t$, we estimate a depth map $D_t$, camera intrinsics $K_t$, and camera-to-world pose $T_t^{c2w} \in \mathrm{SE}(3)$. Each valid pixel $(u,v)$ is back-projected into the world coordinate system as
\begin{equation}
\mathbf{p}_{\mathrm{world}}
=
T_t^{c2w}
\left(
D_t(u,v) K_t^{-1} [u,v,1]^\top
\right).
\end{equation}
Aggregating all back-projected pixels yields a colored point cloud
$\mathcal{P}=\{(\mathbf{x}_i,\mathbf{c}_i)\}$, where
$\mathbf{x}_i \in \mathbb{R}^3$ denotes the 3D position and
$\mathbf{c}_i \in [0,1]^3$ denotes the RGB color. We apply voxel downsampling and statistical outlier removal to reduce noise.

Since monocular reconstruction is scale-ambiguous, we normalize each point cloud to a unit bounding box centered at the origin. The generated point cloud is then aligned to the reference point cloud using FPFH-based global registration followed by point-to-plane ICP refinement.

After alignment, we compute a geometric F-score between the reference point cloud $\mathcal{A}$ and the generated point cloud $\mathcal{B}$. With a distance threshold $\tau$, precision and recall are defined as
\begin{equation}
\mathrm{Prec}(\tau)
=
\frac{1}{|\mathcal{A}|}
\sum_{\mathbf{a}\in\mathcal{A}}
\mathds{1}
\left[
\min_{\mathbf{b}\in\mathcal{B}}
\|\mathbf{a}-\mathbf{b}\|_2 < \tau
\right],
\end{equation}
\begin{equation}
\mathrm{Rec}(\tau)
=
\frac{1}{|\mathcal{B}|}
\sum_{\mathbf{b}\in\mathcal{B}}
\mathds{1}
\left[
\min_{\mathbf{a}\in\mathcal{A}}
\|\mathbf{b}-\mathbf{a}\|_2 < \tau
\right].
\end{equation}
The geometric consistency score is then
\begin{equation}
\mathrm{F\text{-}score}(\tau)
=
\frac{
2\,\mathrm{Prec}(\tau)\,\mathrm{Rec}(\tau)
}{
\mathrm{Prec}(\tau)+\mathrm{Rec}(\tau)
}.
\end{equation}
We set $\tau=0.05$, corresponding to $5\%$ of the normalized bounding-box extent.

To further evaluate appearance consistency, we compute color similarity over geometrically matched point pairs. For each point $\mathbf{b}_j\in\mathcal{B}$, we find its nearest neighbor $\mathbf{a}_{nn(j)}\in\mathcal{A}$ and keep only pairs within the distance threshold:
\begin{equation}
\mathcal{M}
=
\{(j,nn(j)) \mid
\|\mathbf{b}_j-\mathbf{a}_{nn(j)}\|_2 < \tau
\}.
\end{equation}
The color similarity is defined as
\begin{equation}
\mathrm{ColorSim}
=
\frac{1}{|\mathcal{M}|}
\sum_{(j,k)\in\mathcal{M}}
\left(
1 -
\frac{\|\mathbf{c}_j^b-\mathbf{c}_k^a\|_2}{\sqrt{3}}
\right),
\end{equation}
where $\sqrt{3}$ is the maximum possible RGB distance and normalizes the score to $[0,1]$.

Finally, CVC combines geometric and appearance consistency:
\begin{equation}
\mathrm{CVC}
=
\frac{1}{2}
\left(
\mathrm{F\text{-}score}(\tau)
+
\mathrm{ColorSim}
\right).
\end{equation}

\section{Dataset Construction}
\label{sec:dataset construction}

To systematically instantiate the MWSU training tuple from raw single-view videos, we employ an automated, geometry-aware data engine. The construction pipeline consists of explicit feature extraction, spatial trajectory formulation, and environment modeling.

\textbf{Feature extraction and identity formulation.} 
For a given monocular video, we first extract the continuous camera motion sequence $\mathbf{c}_i$ using MoGe-2~\cite{wang2025moge}. Simultaneously, we estimate dense per-frame depth maps using Depth Anything~\cite{yang2024depth} and track dynamic subjects to obtain continuous instance masks via SAM~2~\cite{ravi2024sam}. To formulate the identity set $\mathcal{I}_{\sim i}$, we scan the tracking results for each individual subject $j$ to select a single high-quality frame with a valid, non-empty mask. The foreground is then explicitly cropped to serve as the visual identity reference image $\mathbf{I}_j$.

\textbf{Trajectory abstraction via 3D Gaussians.} 
By aggregating the camera poses, depth maps, and continuous subject masks, we back-project the masked subject regions into global 3D space. Instead of modeling complex, non-rigid meshes, we abstract the subject's spatial geometry at each frame by calculating the mean position and covariance of its point cloud. This efficiently represents the subject as a compact, moving 3D Gaussian sphere. Projecting these global spatial states back onto the camera operator's perspective yields the relative trajectory set $\mathcal{T}_{\sim i}^{(i)}$.

\textbf{Background modeling strategies.}
As described in the main paper, we construct the shared 3D background scene $\mathcal{S}_i$ using two complementary strategies to supervise different generative capabilities. 

\textit{Anchor-Frame Background Modeling} utilizes only the initial frame to establish a partial environmental prior. We remove the subject region using the SAM~2 mask and back-project the remaining background pixels into 3D space to obtain a sparse scene point cloud. This deliberately incomplete prior forces the model to learn scene completion and spatial extrapolation, which is critical for the partial-scene evaluation setting. 

\textit{Holistic Background Aggregation} replaces the single-frame reconstruction with multi-frame background fusion. By applying SAM~2 masks to explicitly filter out dynamic subjects across the entire video, we fuse the remaining per-frame background point clouds into a unified, comprehensive 3D scene point cloud. This dense prior explicitly trains the model to maintain strict spatial consistency and structural fidelity, corresponding to the full-scene evaluation setting.

Finally, the constructed 3D background point cloud and the dynamic Gaussian subject trajectories are jointly rendered along the camera trajectory $\mathbf{c}_i$ to produce a unified condition video. At inference time, users can flexibly choose either background strategy depending on the availability of environmental priors.

\begin{figure}[t]
  \centering
  \includegraphics[width=\linewidth]{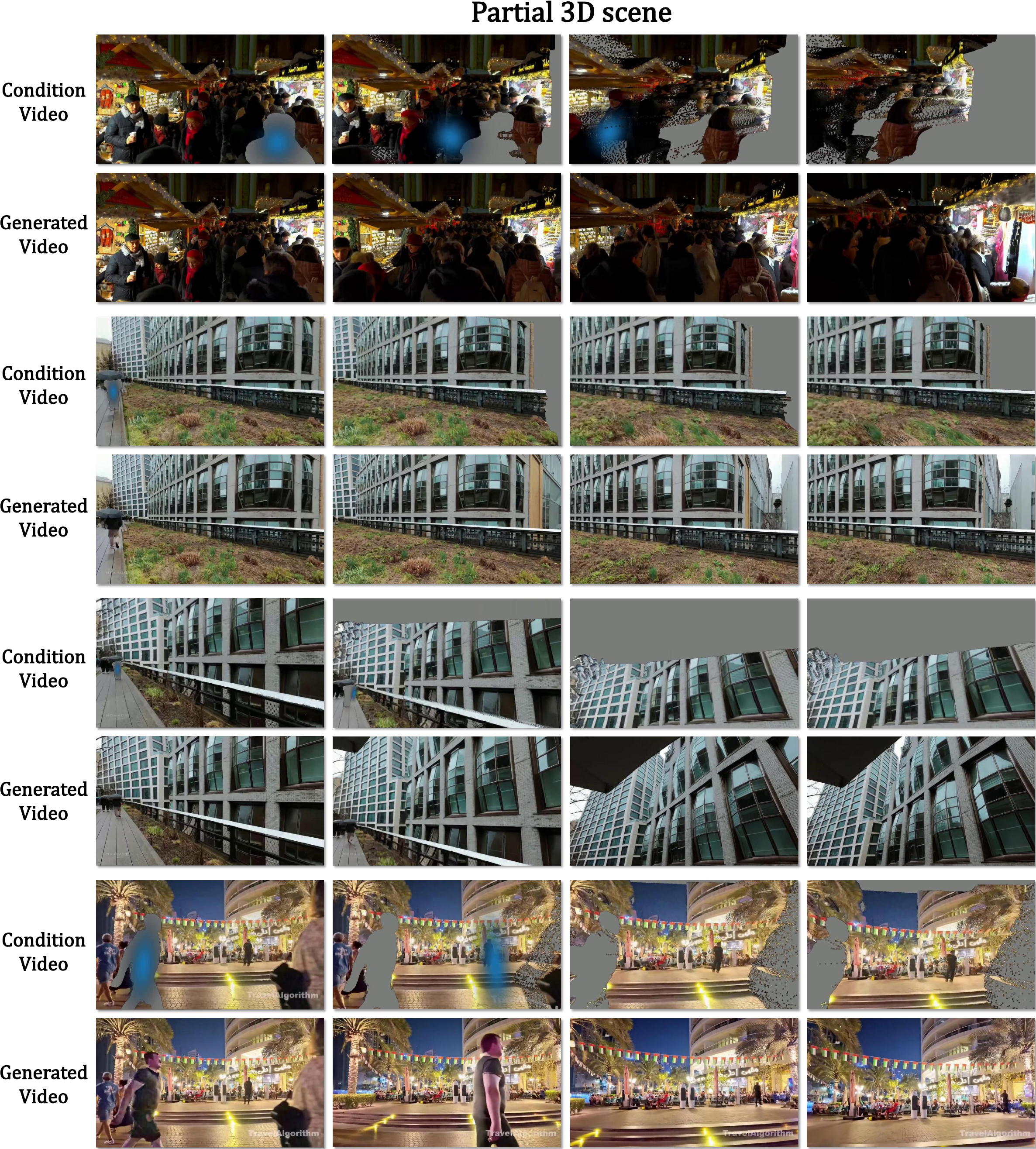}
  \caption{Qualitative single-world generation results on partial 3D scene. \method \ naturally extrapolates and hallucinates unobserved regions as the camera explores.}
  \label{fig:single_partial}
\end{figure}

\begin{figure}[t]
  \centering
  \includegraphics[width=\linewidth]{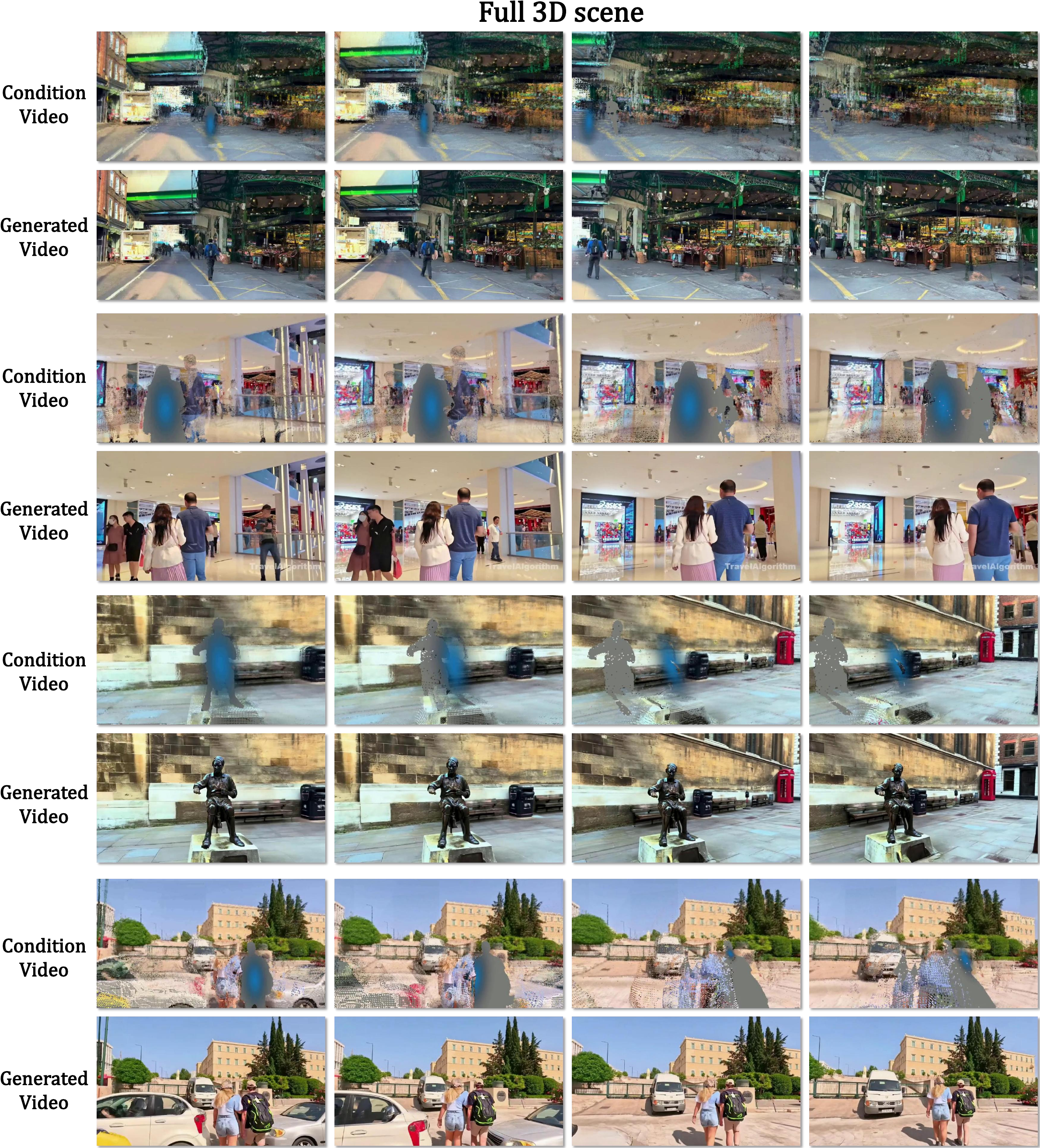}
  \caption{Qualitative single-world generation results on the full 3D scene. \method \  maintains strict spatial consistency and structural fidelity even under large camera rotations.}
  \label{fig:single_full}
\end{figure}

\section{Single-World Generation Results}
\label{sec:suppl_single}

Beyond multi-agent synchronization, \method\ demonstrates robust capabilities as a highly controllable single-view video world model. Our framework grants users fine-grained control over the simulation dynamics: users can explicitly dictate the ego-camera motion as well as the precise 3D spatial trajectories of any objects or subjects within the scene. Furthermore, \method\ supports a free-roaming mode; if no explicit object trajectories are provided, the model seamlessly generates a static environmental walkthrough, allowing the camera to smoothly explore the world without dynamic interruptions.

We demonstrate these capabilities under two distinct scene modeling conditions, reflecting the flexibility of our background representation:

\textbf{Full-scene setting.}
As illustrated in Fig.~\ref{fig:single_full}, when a complete, holistic 3D scene layout is available, \method\ adheres strictly to the comprehensive spatial priors. Conditioned on the dense 3D background point cloud rendered along the camera trajectory, the model preserves exact geometric consistency and coherent spatial structures throughout the video. This holds true even when the camera executes complex trajectories with large rotations (e.g., turning corners or pivoting more than 90 degrees), successfully preventing background shifting or spatial degradation.

\textbf{Partial-scene setting.}
\method\ is equally effective when only limited environmental information is available, as demonstrated in Fig.~\ref{fig:single_partial}. In the partial-scene setting, the background prior is initialized from just a single anchor frame. As the camera explores beyond the initial field of view, the model leverages its powerful pretrained generative priors to naturally hallucinate and extrapolate the unobserved regions. This allows the model to "fill in the blanks" on the fly, producing visually plausible and structurally coherent scene completions without requiring a fully scanned environment.

\begin{figure}[t]
  \centering
  \includegraphics[width=\linewidth]{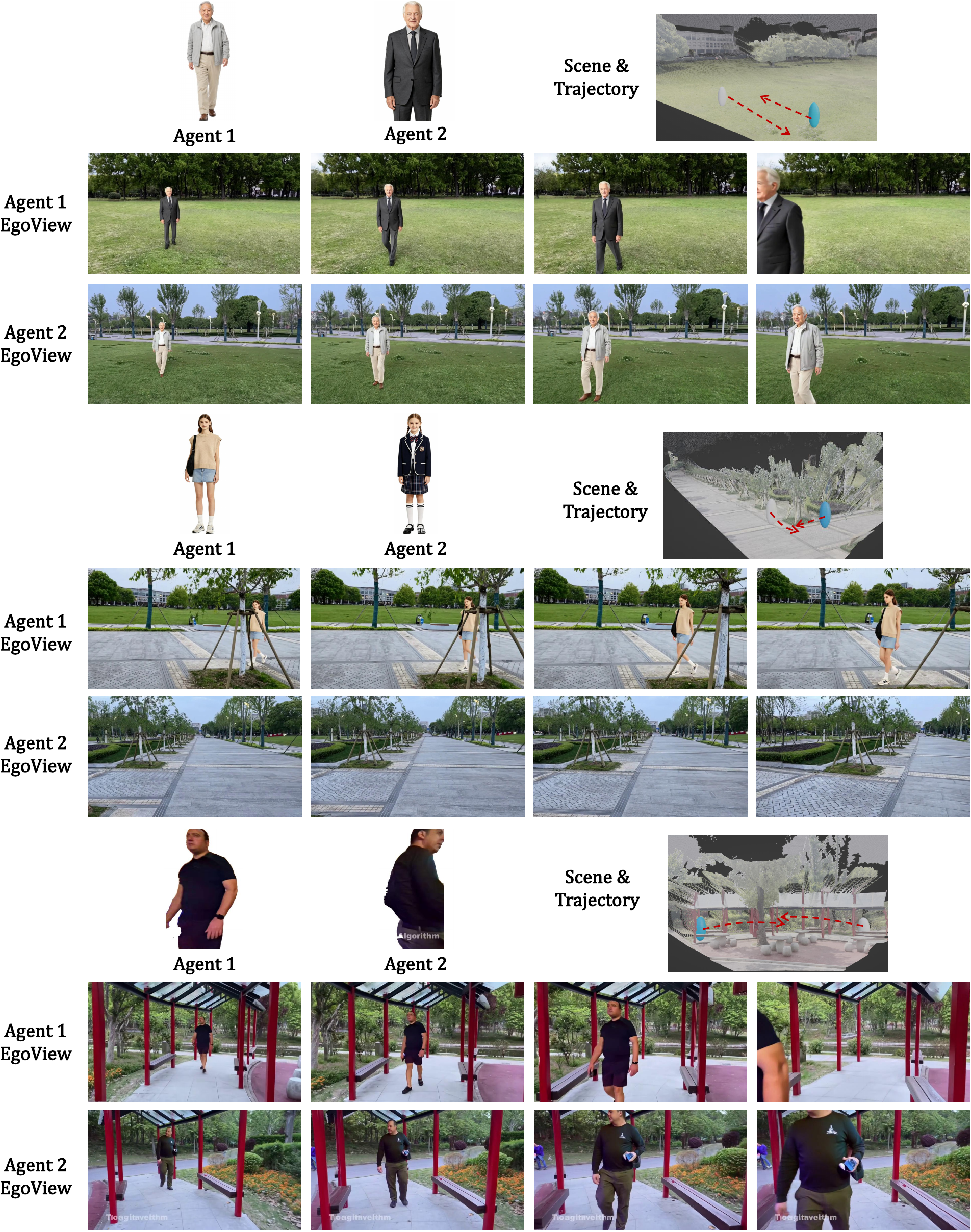}
  \caption{Additional multi-agent generation results.}
  \label{fig:more_results}
\end{figure}

\begin{figure}[t]
  \centering
  \includegraphics[width=\linewidth]{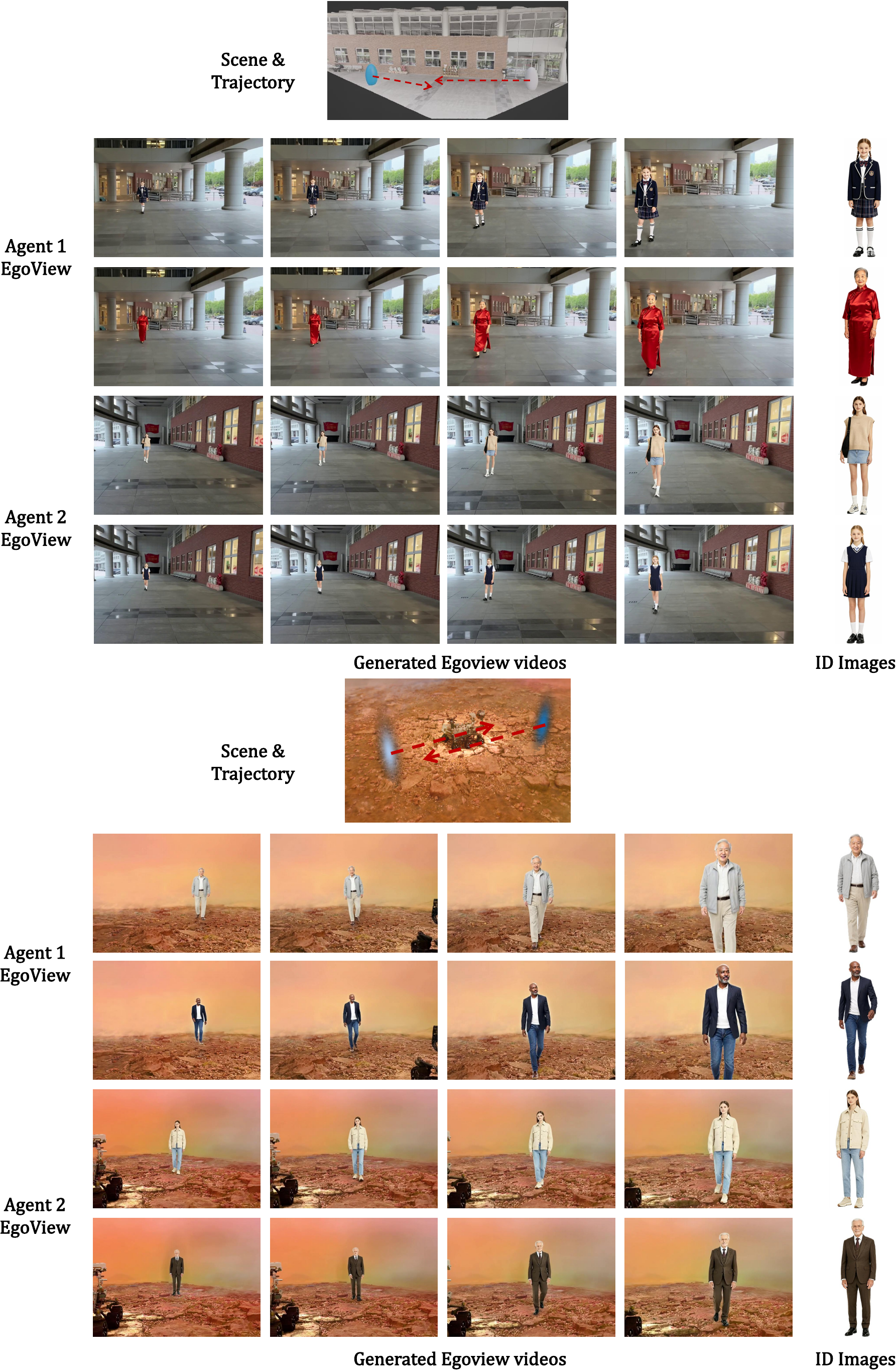}
  \caption{Additional multi-agent generation results. The blocks illustrate diverse open-domain environments and demonstrate character customization, where different specific identities can be seamlessly assigned to agents exploring the exact same shared 3D scene.}
  \label{fig:more_compare}
\end{figure}

\section{Additional Multi-Agent Generation Results}
\label{sec:suppl_more_compare}

To further demonstrate the versatility and robustness of \method, we provide additional qualitative results for the multi-agent generation task, focusing on character customization and scene diversity. 

\textbf{Zero-shot character customization in a shared scene.}
Crucially, \method\ supports highly flexible, zero-shot character customization. As illustrated in Fig.~\ref{fig:more_compare}, because appearance is persistently anchored via the identity reference image, users can explicitly inject customized roles into the simulation. We showcase examples where, within the exact same background scene and adhering to identical spatial trajectories, the visual identities of the agents are seamlessly swapped simply by providing different ID images. This highlights our model's ability to completely disentangle appearance from motion and environment, enabling diverse roleplay and personalized avatars in shared digital realities.

\textbf{Diverse open-domain environments.}
Furthermore, as shown in Fig.~\ref{fig:more_results}, our generated cases span a wide variety of diverse open-domain environments, including complex indoor layouts, expansive outdoor landscapes, and dynamic, unstructured scenes. Across all these categories, \method\ consistently generates physically synchronized and geometrically coherent multi-view egocentric videos, ensuring that both static structures and transient events align perfectly between the interacting agents.





\section{Limitations}
\label{sec:suppl_limitations}

\textbf{World interaction.}
\method\ focuses on open-domain multi-agent world modeling, where multiple agents simultaneously observe a shared physical scene with consistent egocentric views. The current framework does not yet model active interaction between agents and the environment, such as manipulating objects or physically contacting other agents. Extending the system toward interaction-aware dynamics is a natural and interesting direction for future work.

\textbf{Agent orientation.}
Each agent's facing direction is currently derived from the instantaneous velocity of its 3D Gaussian sphere, which works well for typical locomotion trajectories. This means in-place rotation is not yet supported as a trajectory primitive. Adding an explicit heading angle to the trajectory representation would be a straightforward way to enable this.

\textbf{Broader impact.}
\method\ enables realistic multi-agent video simulation for embodied AI research, supporting safe robot learning, game simulation, and social behavior modeling. The ability to generate consistent multi-view videos could be misused for creating synthetic multi-camera surveillance scenarios; we recommend access controls for downstream deployments.

\end{document}